\documentclass[letterpaper, 10 pt, conference]{ieeeconf}  

\IEEEoverridecommandlockouts                      

\usepackage{epsfig} 
\usepackage{amsmath} 
\usepackage{amssymb}  
\usepackage{color}
\usepackage[linesnumbered,ruled]{algorithm2e}
\usepackage[normalem]{ulem} 
\makeatletter
\let\NAT@parse\undefined
\makeatother
\usepackage{hyperref}
\hypersetup{pdfstartview=FitH,
            colorlinks=true,
            linkcolor=red,
            anchorcolor=blue,
            citecolor=green
            }
\usepackage{cite}
\usepackage{booktabs}
\usepackage{multirow}
\usepackage[caption=false,font=footnotesize]{subfig}
\usepackage[dvipsnames]{xcolor}
\usepackage{subfig}
\usepackage{soul}
\usepackage{balance}
\newcounter{RNum}
\renewcommand{\theRNum}{\arabic{RNum}}
\newcommand{\Remark}{\noindent\textit{\textbf{Remark}~\refstepcounter{RNum}\textbf{\theRNum}: }}
\newcommand{\NoOne}[1]{\textcolor{red}{#1}}
\newcommand{\NoTwo}[1]{\textcolor{green}{#1}}
\newcommand{\NoThree}[1]{\textcolor{blue}{#1}}
\soulregister\NoOne7
\soulregister\NoTwo7
\soulregister\NoThree7
\soulregister\Remark7

\title{\LARGE \bf
SAM-DA: UAV Tracks Anything at Night with SAM-Powered\\Domain Adaptation
}

\author{Changhong Fu$^{1,*}$, Liangliang Yao$^{1}$, Haobo Zuo$^{2}$, Guangze Zheng$^{2}$, Jia Pan$^{2}$
\thanks{$^{*}$Corresponding author}%
\thanks{$^{1}$Changhong Fu and Liangliang Yao are with the School of Mechanical Engineering, Tongji University, Shanghai 201804, China.
        {\tt\footnotesize E-mail: changhongfu@tongji.edu.cn}}%
\thanks{$^{2}$Haobo Zuo, Guangze Zheng, and Jia Pan are with the Department of Computer Science, University of Hong Kong, Hong Kong 999077, China.}
}

\begin{document}

\maketitle
\thispagestyle{empty}
\pagestyle{empty}


\begin{abstract}
Domain adaptation (DA) has demonstrated significant promise for real-time nighttime unmanned aerial vehicle (UAV) tracking. However, the state-of-the-art (SOTA) DA still lacks the potential object with accurate pixel-level location and boundary to generate the high-quality target domain training sample. This key issue constrains the transfer learning of the real-time daytime SOTA trackers for challenging nighttime UAV tracking. Recently, the notable Segment Anything Model (SAM) has achieved a remarkable zero-shot generalization ability to discover abundant potential objects due to its huge data-driven training approach. To solve the aforementioned issue, this work proposes a novel SAM-powered DA framework for real-time nighttime UAV tracking, \textit{i.e.}, SAM-DA. Specifically, an innovative SAM-powered target domain training sample swelling is designed to determine enormous high-quality target domain training samples from every single raw nighttime image. This novel one-to-many generation significantly expands the high-quality target domain training sample for DA. Comprehensive experiments on extensive nighttime UAV videos prove the robustness and domain adaptability of SAM-DA for nighttime UAV tracking. Especially, compared to the SOTA DA, SAM-DA can achieve better performance with fewer raw nighttime images, \textit{i.e.}, the fewer-better training. This economized training approach facilitates the quick validation and deployment of algorithms for UAVs. The code is available at \url{https://github.com/vision4robotics/SAM-DA.}
\end{abstract}

\section{Introduction} \label{sec:intro}

Object tracking has been applied for wide unmanned aerial vehicle (UAV) applications, \textit{e.g.}, geographical research \cite{8880656}, dynamic object investigation \cite{8736008}, search and rescue mission~\cite{varga2022seadronessee}. 
Using abundant daytime superior-quality tracking datasets \cite{8922619,lin2014microsoft}, state-of-the-art (SOTA) trackers~\cite{cao2021hift,9577739,8953931,9991169} have attained remarkable performance. Nevertheless, the performance of these SOTA trackers is unsatisfactory in darkness due to the limited illumination, low contrast, and much noise of nighttime images in comparison to daytime ones~\cite{9696362,9636680}. The aforementioned distinctions bring the discrepancy in feature distribution between day and night images. A potential solution is capturing and annotating sufficient nighttime data for directly training effective nighttime trackers. However, it is expensive and time-consuming to label a large amount of high-quality tracking data under unfavorable lighting conditions. 
\begin{figure}[t!]	
	\centering
	\includegraphics[width=0.95\linewidth]{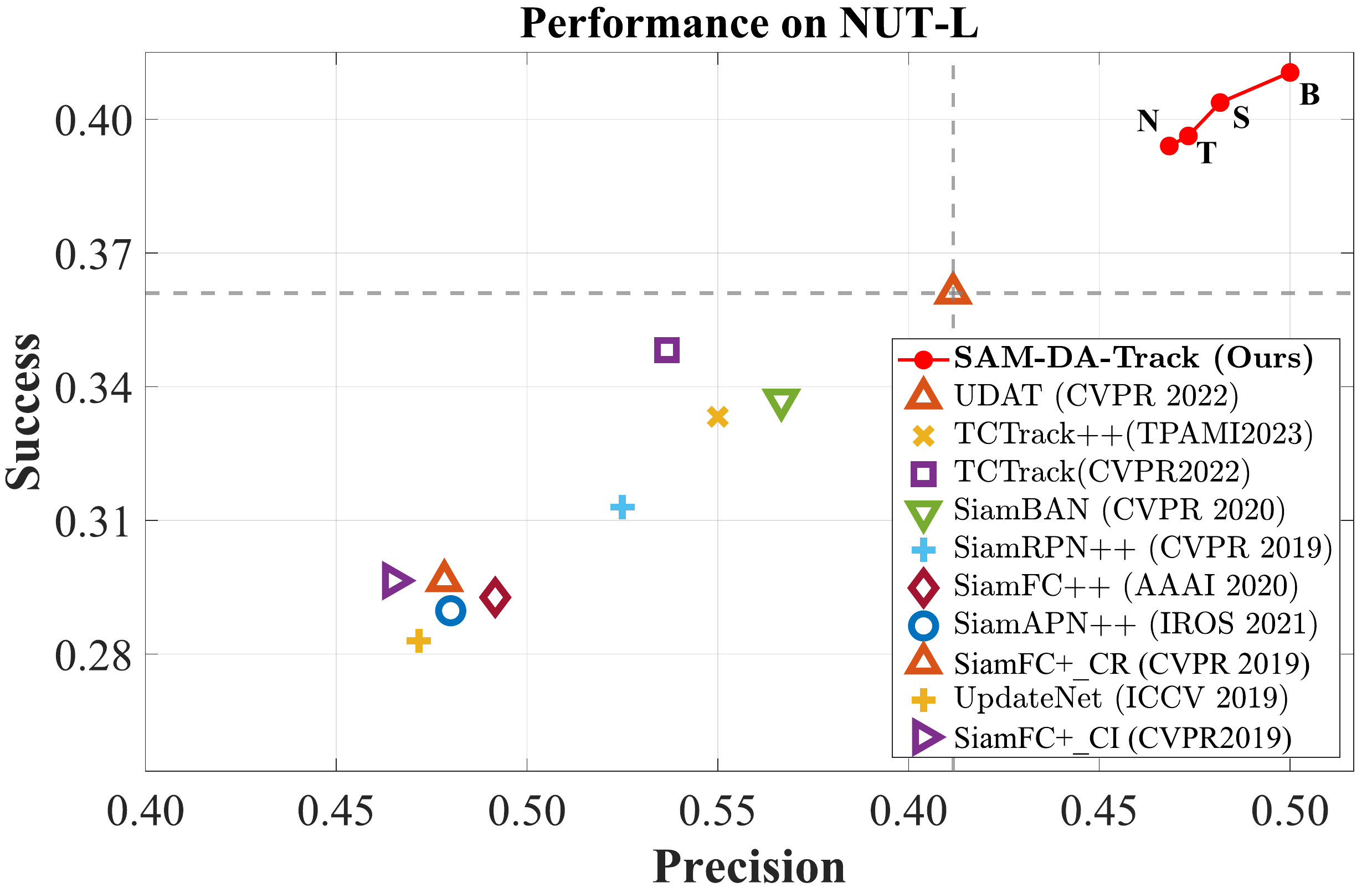}
        \vspace{-2pt}
	\caption
    {
	Overall performance of SAM-DA-Track and state-of-the-art (SOTA) trackers on the proposed NUT-L. SAM-DA-Track represents the version of the base tracker SiamBAN~\cite{9157457}, adopting SAM-DA for adaptation training. N, T, S, and B represent that the target domain training images of SAM-DA-Track are about 10.0\%, 33.2\%, 50.1\%, and 100\% of the entire NAT2021-$train$~\cite{9879981}, respectively. With SAM, the proposed SAM-DA-Track shows superior performance with only 10\% of training images of Baseline UDAT~\cite{9879981}. 
	}
	\label{fig1}
 \vspace{-3pt}
\end{figure}

Considering the labeling cost of the nighttime image and the domain gap of day-night, domain adaptation \cite{9879981} is introduced to solve the problem of nighttime UAV tracking. This method aims to transfer SOTA trackers developed for daytime situations to nighttime UAV tracking. In the source domain, the training data has well-annotated bounding boxes with high expenses by hand, whereas, in the target domain, the training samples are obtained by the automatic method \cite{zhang2021dynamic} instead of manual annotation. However, the insufficient quality of target domain training samples limits the improvement of domain adaptation. the obtaining of training samples has difficulty discovering the potential object with precise pixel-level location and boundary from challenging nighttime images of UAV perspectives~\cite{9320524}. Furthermore, this method solely concentrates on one target domain training sample within a single nighttime image, disregarding abundant other valuable potential objects, \textit{i.e.}, one-to-one generation. \textit{Therefore, how to generate enormous and high-quality target domain training samples from every single raw nighttime image for robust day-night domain adaptation is an urgent problem.}

Recently, the Segment Anything Model (SAM) \cite{kirillov2023segment} has demonstrated an impressive zero-shot generalization capacity, offering greater potential to discover numerous and diverse objects. This achievement can be attributed to its huge data-driven training approach with over one billion masks. 
Such kind of generalization ability enables SAM to be directly applied for various vision-based tasks without task-oriented training, including camouflaged object detection~\cite{tang2023can}, medical image segmentation~\cite{roy2023sam}, \textit{etc}. Moreover, with the enormous parameters, SAM is capable of extracting robust image features in various environments. Despite the above advantages, SAM is hard to be directly applied for nighttime UAV tracking due to the limited load source and computation power of the UAV. \textit{Thereby, how to effectively utilize the considerable zero-shot generalization ability of SAM for real-time nighttime UAV tracking is worth exploring carefully.}

This work introduces the superb SAM into the training phase of tracking-oriented day-night domain adaptation for the first time, proposing a novel SAM-powered domain adaptation framework, \textit{i.e.}, SAM-DA. Specifically, the inventive SAM-powered target domain training sample swelling is presented to determine enormous high-quality target domain training samples from every single challenging nighttime image, dubbed the one-to-many generation. Thereby, the dependence on the number of raw images required for adaptation training can be reduced to enhance generalization and prevent overfitting. With the improvement and increase of target domain training samples, the adaptation effect of SOTA trackers for nighttime UAV tracking can be further boosted. Figure~\ref{fig1} shows the tracking performance comparison of SAM-DA-Track and other SOTA tracking methods on a comprehensive long-term nighttime UAV tracking benchmark, \textit{i.e.}, NUT-L, which is a combination of long-term sequences from NAT2021-$test$~\cite{9879981} and UAVDark135~\cite{li2022all}. SAM-DA-Track symbolizes the version of the base tracker, \textit{i.e.}, SiamBAN \cite{9157457}, using SAM-DA for adaptation training. N, T, S, B represent that the target domain training images of SAM-DA-Track are about 10.0\%, 33.2\%, 50.1\%, and 100\% of the entire NAT2021-$train$~\cite{9879981}, respectively. The Baseline is UDAT~\cite{9879981}. Compared to this method, SAM-DA-Track adopting the training framework SAM-DA can achieve superior tracking performance with less raw nighttime images, \textit{i.e.}, the few-better training.
The main contributions of this work are as follows:
\begin{itemize}
\item A novel SAM-powered domain adaptation framework, namely SAM-DA, is proposed for real-time nighttime UAV tracking. According to our knowledge, SAM-DA is the first work to combine SAM with domain adaptation for UAV tracking at night.
\item An innovative SAM-powered target domain training sample swelling is designed to determine enormous high-quality target domain training samples from every single raw nighttime image.
\item Comprehensive experiments on extensive nighttime videos verify the effectiveness and domain adaptability of SAM-DA for nighttime UAV tracking. Especially, SAM-DA realizes better performance with fewer raw images compared to the SOTA method. The above training approach promotes the quick validation and deployment of algorithms for UAVs.
\end{itemize}

\section{Related work}
\subsection{Nighttime UAV tracking} 

Recently, nighttime UAV tracking has been utilized in a variety of practical applications, attracting widespread interest. 
Initially, the tracking-oriented low-light enhancers~\cite{9696362,9636680} are designed to improve the nighttime tracking performance of the cutting-edge Siamese trackers~\cite{8954116,9636309,yao2023sgdvit}. Specifically, J. Ye \textit{et al.} \cite{9636680} develop an enhancer to iteratively mitigate the effects of inadequate illumination and noise. Afterward, they present a spatial-channel Transformer-based low-light enhancer to achieve robust nighttime UAV tracking \cite{9696362}. Nevertheless, this plug-and-play method has a restricted relationship with tracking tasks, and the way to directly insert tracking models is unable to minimize the image feature distribution gap. Besides, the model parameters of the low-light enhancers will seriously increase the burden of the limited UAV computation and resources.
        \begin{figure*}[!t]	
		\centering
		\includegraphics[width=0.97\linewidth]{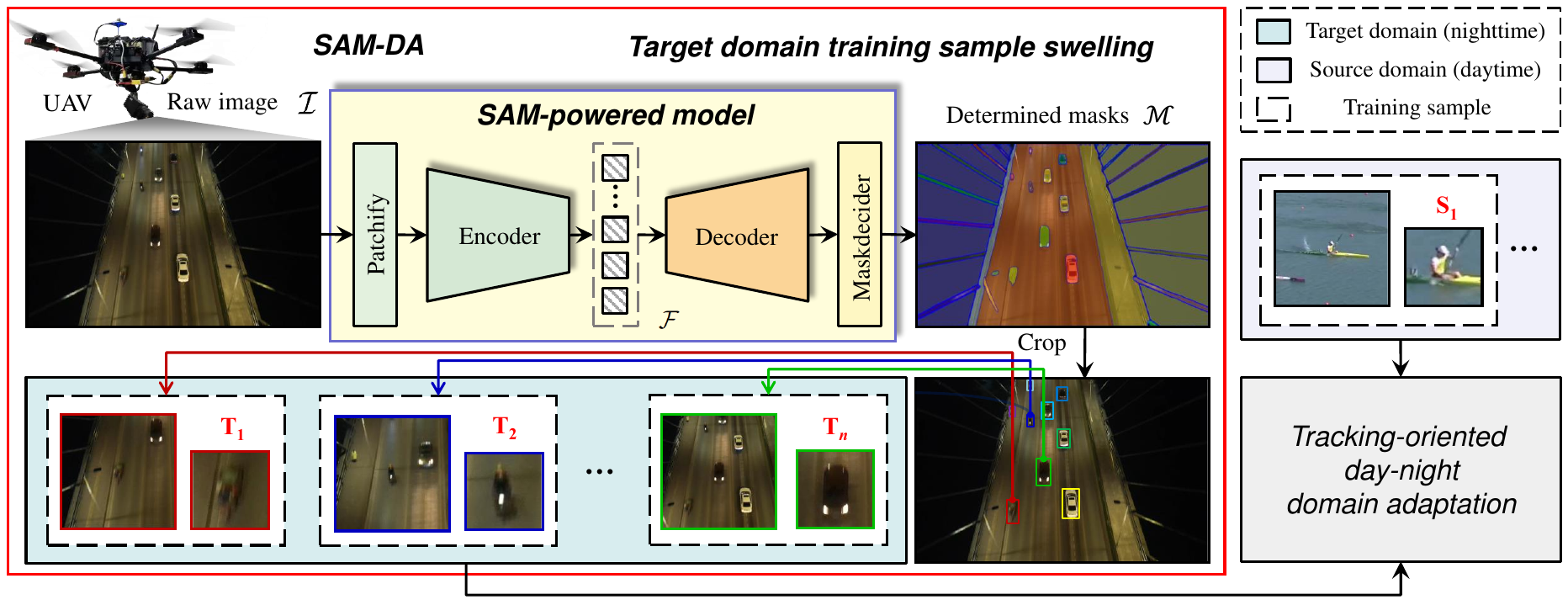}
			\setlength{\abovecaptionskip}{3pt}
		\caption
		{Illustration of the proposed SAM-DA for nighttime UAV tracking. The original nighttime image is from NAT2021-$train$. The source domain training sample is from GOT-10k~\cite{8922619}.
          The SAM-powered target domain training sample swelling is employed to determine enormous high-quality target domain training samples from every single nighttime image. 
          Note the source domain (daytime) training samples are manually collected with time-consuming and expensive annotation, while the target domain (nighttime) training samples are automatically generated with our time-saving and low-cost swelling. 
		}  
		\label{fig:main}
		\vspace{-10pt}
	\end{figure*}
\subsection{Day-night domain adaptation}
Day-night domain adaptation has been used for a wide range of visual tasks~\cite{9577711,sasagawa2020yolo} because it can decrease the domain gap and transfer information from the source domain (daytime) to the target domain (nighttime). X. Wu \textit{et al.} \cite{9577711} train a domain adaptation model for semantic segmentation at night using adversarial learning. Y. Sasagawa \textit{et al.} \cite{sasagawa2020yolo} use domain adaptation to combine deep learning models from various disciplines to detect nighttime objects. Despite the rapid development in other vision tasks, day-night domain adaptation still lacks research for object tracking. Thereby, UDAT \cite{9879981} introduces unsupervised domain adaptation into nighttime UAV tracking, thus increasing the tracking performance at night. However, the insufficient quality of target domain training samples constrains the advancement of the domain adaptation performance for nighttime UAV tracking. Due to the challenges of nighttime images from UAV perspectives, the existing generation approach \cite{zhang2021dynamic} of training samples
struggles to extract the potential object with exact pixel-level location and boundary. In addition, this kind of approach only considers one target domain training sample inside a single nighttime image, ignoring abundant additional worthwhile potential objects.

\subsection{Segment anything model}
SAM \cite{kirillov2023segment} has found extensive use in many kinds of computer vision tasks as a huge data-driven method. Trained with over a billion masks, the renowned SAM has extraordinary zero-shot generalization ability. The above capability allows SAM to be directly applied to different vision-based tasks. Specifically, L. Tang \textit{et al.}~\cite{tang2023can} provide an initial assessment for the efficacy of SAM on the camouflaged object detection assignment. S. Roy \textit{et al.} \cite{roy2023sam} seek to undertake an early assessment of the out-of-the-box zero-shot capabilities of SAM for medical image segmentation. Despite its wide applications in other vision tasks, SAM has not been used for UAV tracking, especially in nighttime scenarios.

\section{Proposed method}
SAM-DA is introduced in this section, as depicted in Fig.~\ref{fig:main}. Given a nighttime raw image, the SAM-powered target domain training sample swelling is employed to determine enormous potential objects and provide their accurate pixel-level locations and boundaries. Then, the number of training samples swells from one to many within every single nighttime image based on the above pixel-level locations and boundaries. During the training pipeline, both the manually annotated source domain (daytime) training sample and automatically generated target domain (nighttime) training sample are leveraged to drive the following tracking-oriented day-night domain adaptation. This data-driven framework introduces SAM into the training phase of domain adaptation, improving the performance of the tracker for nighttime UAV tracking.


   
\subsection{SAM-powered target domain training sample swelling}
Effective day-night domain adaptation requires enormous high-quality target domain training samples.   
Different from existing solutions where only one training sample is obtained from one image~\cite{zhang2021dynamic}, the proposed SAM-powered target domain training sample swelling utilizes the powerful zero-shot generalization ability of SAM to swell the number of high-quality training samples. 

\noindent\textbf{SAM-powered model.} As shown in Fig.~\ref{fig:main}, following SAM, a nighttime raw image $\mathcal{I}\in \mathcal{R}^{H\times W \times 3}$ is first patchified to patch embedding. Subsequently, an encoder is utilized to extract feature embeddings denoted by $\mathcal{F}$ as follows:
       \begin{equation}
		\begin{split}
                \mathcal{F} &= \mathrm{Encoder}(\mathrm{Patchify}(\mathcal{I}))\quad,
		\end{split}
	\end{equation}
where $\mathrm{Patchify}$ means that the image is projected linearly and added with position embeddings. $\mathrm{Encoder}$ represents an MAE~\cite{he2022masked} pre-trained Vision Transformer (ViT)~\cite{dosovitskiy2020image}.

Afterward, a decoder predicts the mask embeddings. Then, a mask decider generates the image with enormous determined masks $\mathcal{M}$. The process is defined as:
        \begin{equation}
		\begin{split}
               \mathcal{M} &= \mathrm{Maskdecider}( \mathrm{Decoder}( \mathcal{F} ) )\quad,
		\end{split}
	\end{equation}
where $\mathrm{Decoder}$ is a modification of a Transformer decoder block~\cite{vaswani2017attention} and $\mathrm{Maskdecider}$ is a dynamic mask prediction head. 
These masks contain information about the potential object with accurate location and boundary. Hence, the boxes $[\mathbf{B}_1,\mathbf{B}_2,...,\mathbf{B}_n]$ around the masks are utilized to generate target domain training samples. $n$ is the number of potential objects for a single nighttime image.

\begin{figure}[!t]	
		\centering
		\setlength{\abovecaptionskip}{5pt}
		\includegraphics[width=1\linewidth]{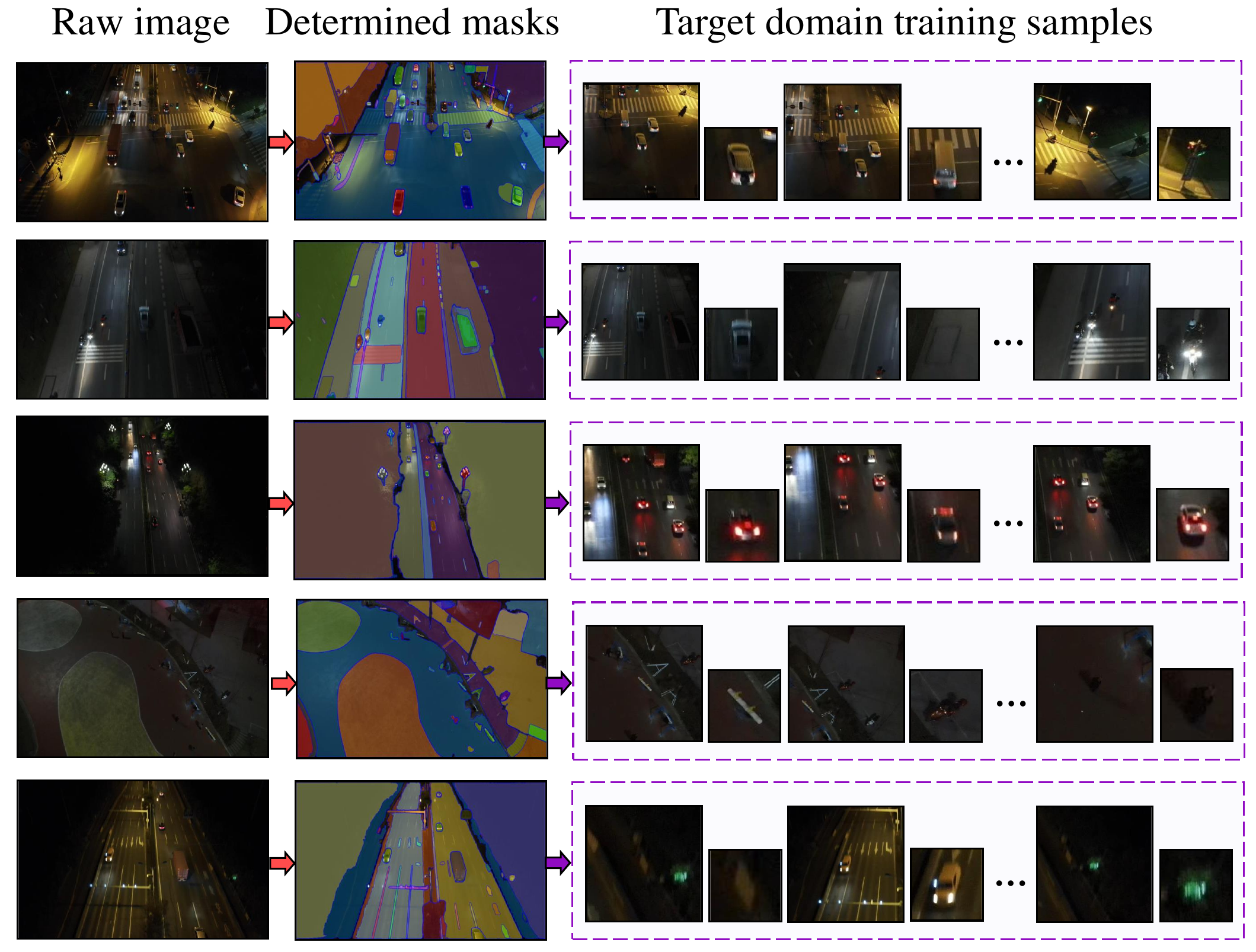}
		\caption
		{Visualization of the target domain training samples processed by the SAM-powered target domain training sample swelling. The nighttime raw images are from NAT2021-$train$. With the novel one-to-many generation, many objects, such as buses, cars, riders, and signals are utilized to generate the training samples from a single image, enhancing the utilization efficiency of the raw nighttime data.   
            }
		\label{fig:visual}
	\end{figure}
\Remark The original SAM is difficult to be utilized straightforwardly for real-time nighttime UAV tracking due to the restricted load source and processing capabilities of the UAV. Therefore, this work adopts SAM  to generate high-quality target domain training samples for day-night domain adaptation training. 
 
\noindent\textbf{Target domain training sample swelling.} The target domain training sample swelling generates many high-quality training samples from one original nighttime image. Specifically, this work follows the data processing of COCO~\cite{lin2014microsoft}. An image $\mathcal{I}$ exhibits the dual functionality of serving as both a template frame and a search frame.
As a template frame, referring to the boxes $[\mathbf{B}_1,\mathbf{B}_2,...,\mathbf{B}_n]$, the image is cropped into numerous target-centered image patches, denoted as template patches $\left[\mathcal{Z}_1,\mathcal{Z}_2,...,\mathcal{Z}_n\right]$, which are subsequently resized to a fixed size (\textit{e.g.}, 127 × 127). Simultaneously, as a search frame, the image is cropped into an equal number of larger image patches, denoted as search patches $\left[\mathcal{X}_1,\mathcal{X}_2,...,\mathcal{X}_n\right]$, also based on the predicted boxes $[\mathbf{B}_1,\mathbf{B}_2,...,\mathbf{B}_n]$, and resized to another size (\textit{e.g.}, 255 × 255). Patches containing the same target are paired between the template patches and search regions patches, forming abundant target domain training samples $[\mathbf{T}_1,\mathbf{T}_2, ...,\mathbf{T}_n]$, where $\mathbf{T}_i=\{\mathcal{Z}_i,\mathcal{X}_i\}$. The target domain training sample swelling is defined as:

        \begin{equation}
		\begin{split}
                &\left[\mathcal{Z}_1, \mathcal{Z}_2,...,\mathcal{Z}_n\right] = \mathrm{Crop}( \mathcal{I}; [\mathbf{B}_1,\mathbf{B}_2,...,\mathbf{B}_n], s_1)\quad,\\
                &\left[\mathcal{X}_1,
                \mathcal{X}_2,...,\mathcal{X}_n\right] = \mathrm{Crop}(\mathcal{I};[\mathbf{B}_1,\mathbf{B}_2,...,\mathbf{B}_n],s_2)\quad,\\
		\end{split}
	\end{equation}
where $\mathrm{Crop}$ is the crop operation to generate patches and $\mathbf{B}_i$ is the box of the i-th potential object. Besides, $s1,s2$ represent template size and search size, respectively.

\Remark As shown in Fig.~\ref{fig:visual}, the visualization represents the swelling method from one original nighttime image to many target domain training samples. The one-to-many generation revolutionizes the existing target domain
training sample acquisition approach in the day-night
domain adaptation from both the aspects of quantity and
quality.

\subsection{Tracking-oriented day-night domain adaptation}
With a substantial amount of high-quality target domain training samples, a tracking-oriented day-night domain adaption is utilized to enhance the trackers’ nighttime performance by aligning the features from both the source domain (daytime) and the target domain (nighttime). Following the paradigm of Baseline, the whole domain adaptation framework is divided into four parts: backbone, bridging module, tracker head, and discriminator.
 
\noindent\textbf{Backbone.}
	In a general Siamese network-based tracker, feature extraction involves two branches, the template branch and the search branch.
    These branches utilize an identical backbone network to generate feature maps from the template patch and search patch. Since the heavy computation brought by the deep structure hardly be afforded by the UAV platform, the lightweight convolutional neural network (CNN)-based network is usually adopted.
 
\noindent\textbf{Bridging module.}
In consideration of the domain discrepancy between the source domain (daytime) and the target domain (nighttime), the bridging module is designed to bridge the gap between the feature distributions. The Transformer-based network is the preferred choice due to the strong modeling capability for long-range inter-independencies.

\noindent\textbf{Tracker head.} After the bridging module, a cross-correlation operation is calculated on the modulated features to generate a similarity map. Finally, the tracker head performs the classification and regression process based on the similarity map to predict the position of the object.

\noindent\textbf{Discriminator.}
The day-to-night domain adaptation framework is trained via an adversarial learning paradigm. Within this adversarial learning framework, the discriminator is optimized to accurately discern the origin of features, distinguishing between the source and target domains.
\begin{figure}[!t]	
	\centering
{
	\includegraphics[width=0.96\linewidth]{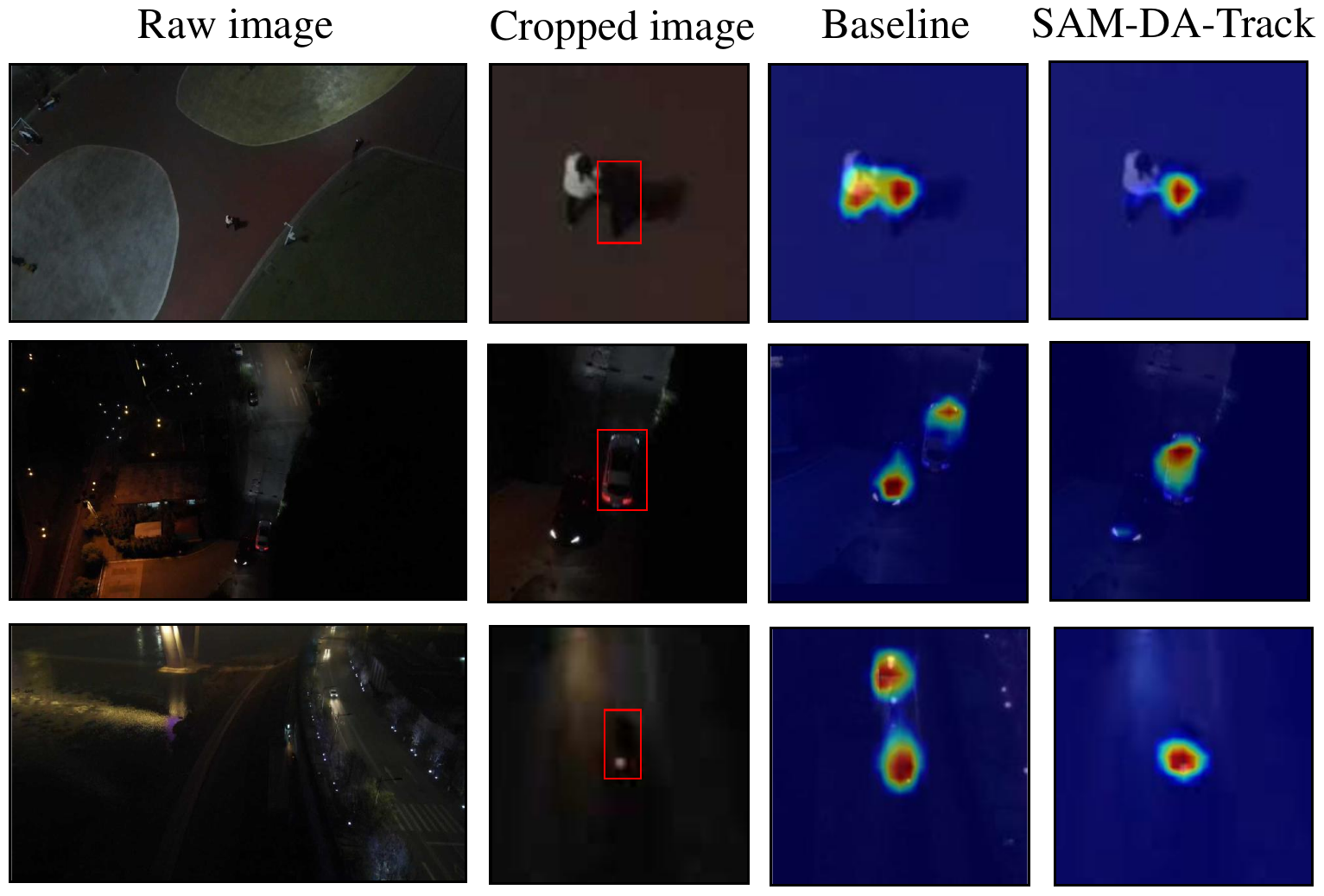}
 }
	\caption
	{
		Visual comparison of confidence maps generated by the Baseline and the SAM-DA-Track. Target objects are marked by \textcolor[rgb]{1,0,0}{red} boxes. The nighttime images are from the proposed NUT-L. The Baseline exhibits sub-optimal performance in tracking tasks conducted under low-light conditions, whereas the proposed SAM-DA-Track method demonstrates notable efficacy in such scenarios.
	}
	\label{heatmap}
\end{figure}


\begin{figure*}[t!]
\subfloat[Precision, normalized precision, and success plots on DarkTrack2021.]{
	\centering
    {\includegraphics[width=0.32\linewidth]{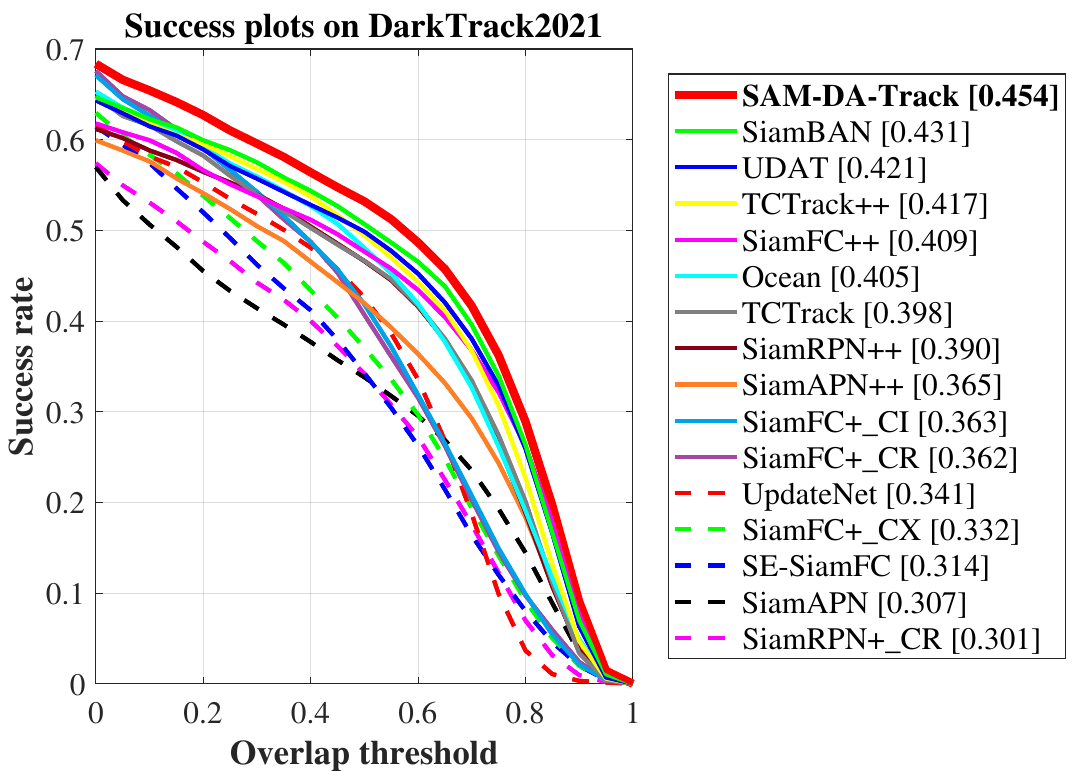}	\includegraphics[width=0.32\linewidth]{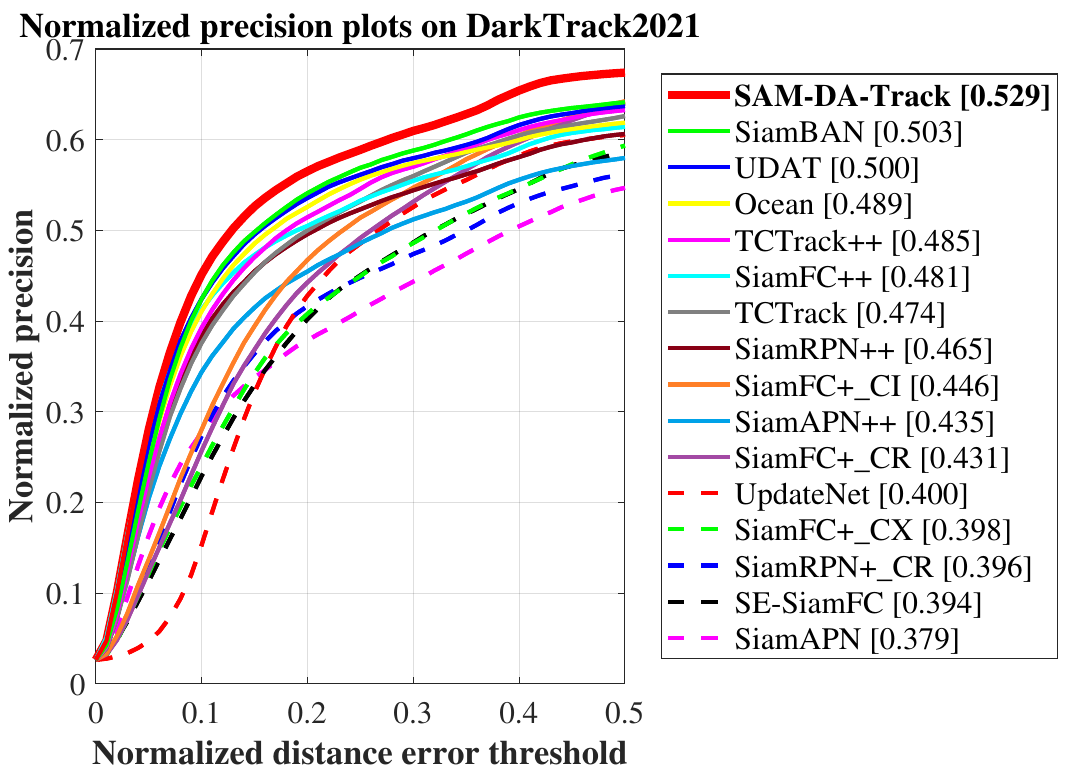}
    \includegraphics[width=0.32\linewidth]{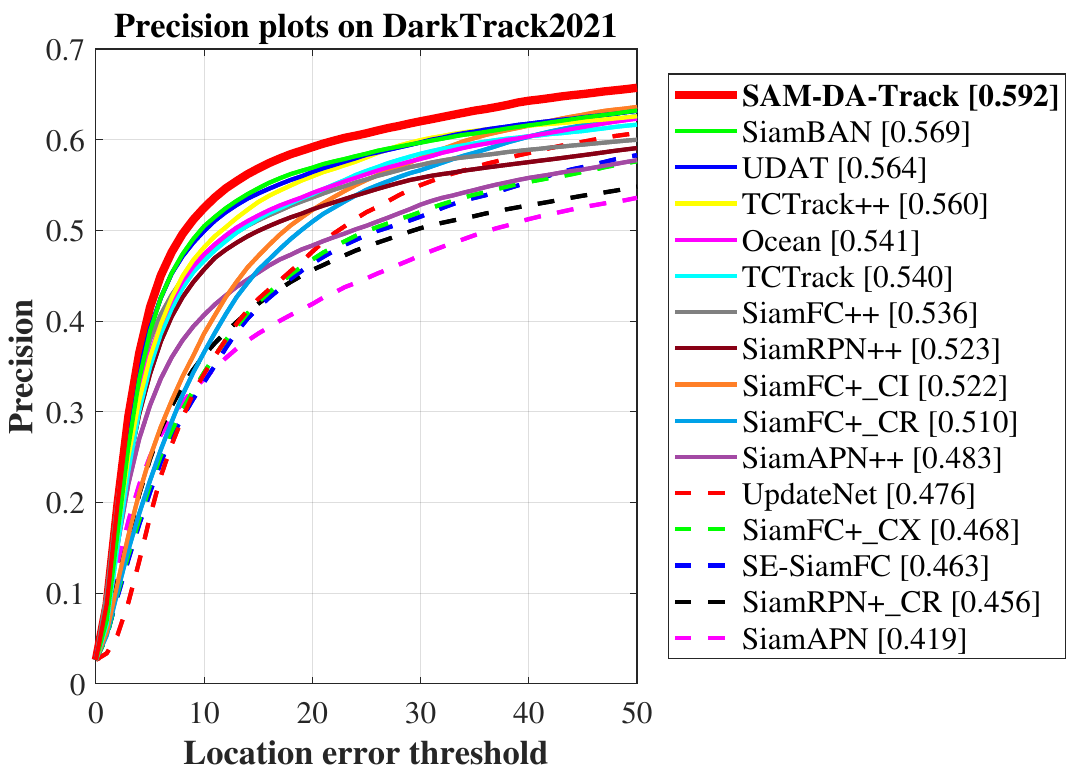}
    }}
    
    \subfloat[Precision, normalized precision, and success plots on NUT-L.]{
	\centering
{\includegraphics[width=0.32\linewidth]{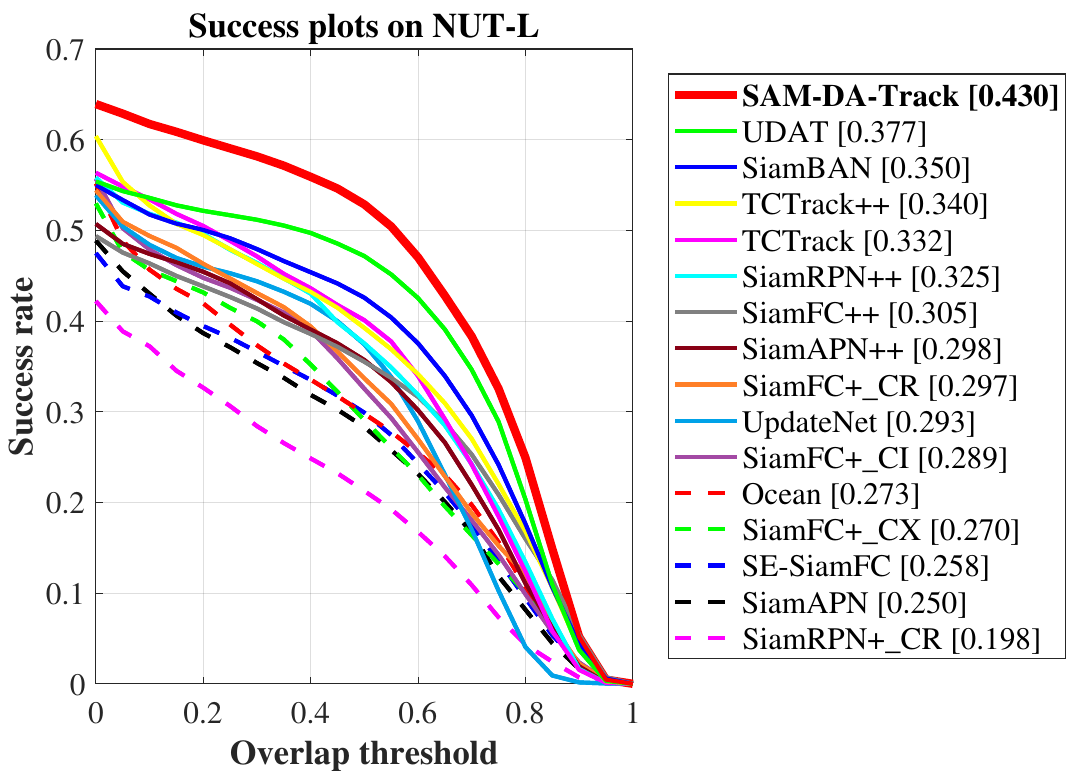}	\includegraphics[width=0.32\linewidth]{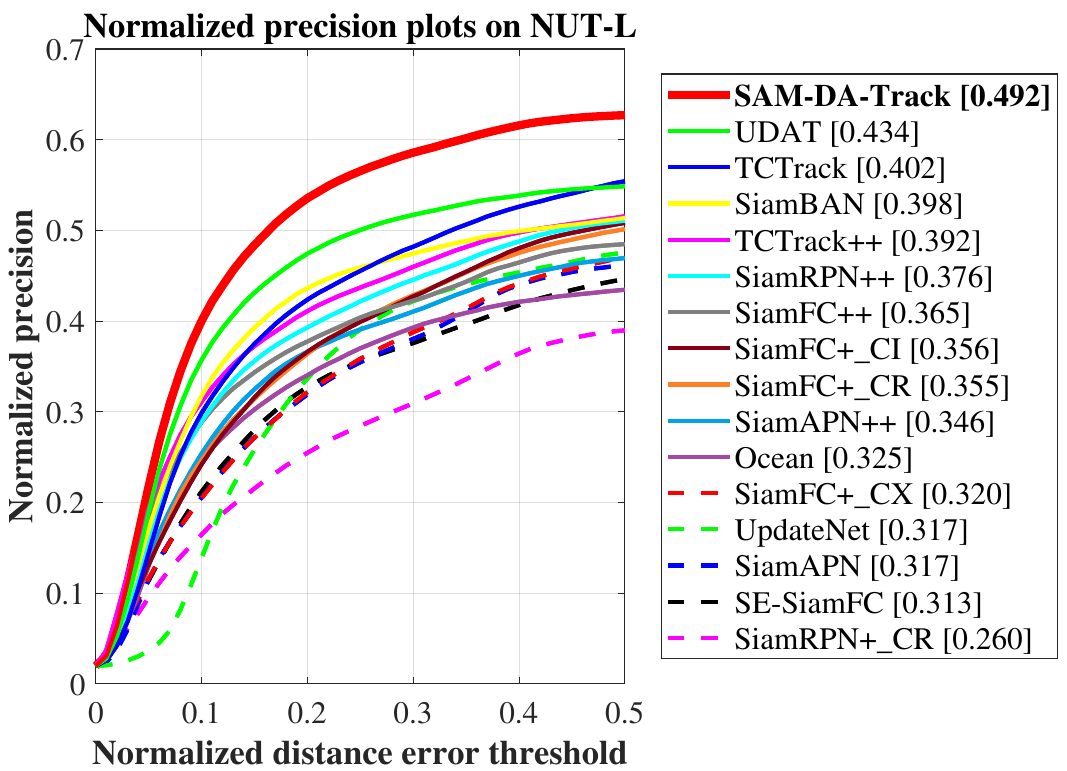}
\includegraphics[width=0.32\linewidth]{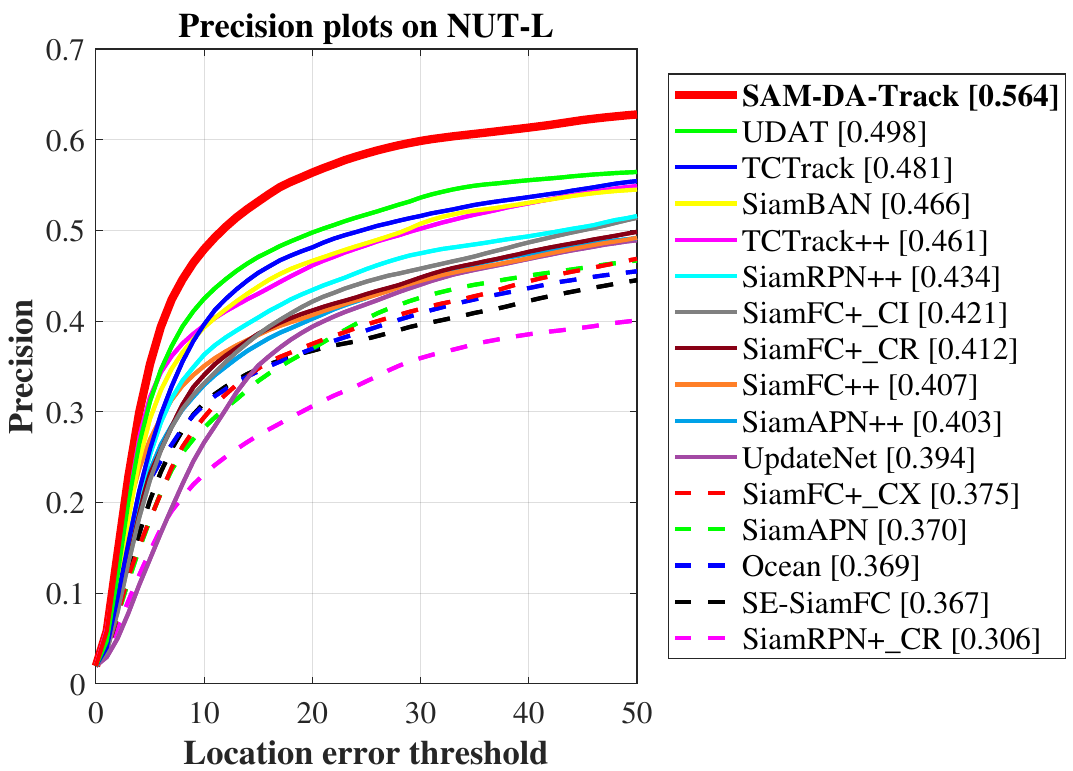}
}}
	\caption
	{
		Overall performance of SAM-DA-Track and SOTA trackers on DarkTrack2021 \cite{9696362} and the proposed NUT-L. SAM-DA-Track significantly surpasses the Baseline and other SOTA methods.
	}
	\label{fig:overall}
 \vspace{-15pt}
\end{figure*}


\Remark Figure~\ref{heatmap} shows the visual comparison of confidence maps generated by the Baseline, and the SAM-DA-Track. With a substantial amount of high-quality target domain training samples, the tracking-oriented day-night domain adaptation greatly improves the performance of the tracker at night.

\section{Experiments}
\subsection{Implementation details}

\noindent\textbf{Data.}
SAM-powered target domain training sample swelling is implemented on NAT2021-$train$ to validate the influence of target domain training samples' quantity and quality on nighttime tracking performance. Four versions of the target domain training set are obtained according to the number of used raw images, \textit{i.e.}, base (B), small (S), tiny (T), and nano (N).
{SAM-NAT-B} swells 100\% original images in the entire NAT2021-$train$, while {SAM-NAT-N}, {SAM-NAT-T}, and {SAM-NAT-S} only randomly sample parts of data with the ratio of about 10.0\%, 33.2\%, and 50.1\%, respectively. 
The quantitative comparison of the training sample numbers between {{SAM-NAT}} and NAT2021-$train$ is shown in Tab.~\ref{tab:all} and discussed in Sec.~\ref{sec:labelswelling} to show the training sample diversity of the proposed SAM-powered swelling. All the training is complemented with PyTorch on a single NVIDIA A100 GPU and follows~\cite{9879981} with the base tracker~\cite{9157457}. No additional data is introduced.

\noindent\textbf{Evaluation.}
To validate the tracking robustness in practical UAV applications against complicated challenges, the performances on SOTA nighttime tracking benchmark DarkTrack2021 \cite{9696362} are analyzed. Furthermore, from the nighttime tracking benchmarks, NAT2021-$test$ and UAVDark135, the long-term tracking videos are combined following the rules in the Baseline~\cite{9879981} and form \textbf{NUT-L}, a comprehensive \textbf{L}ong-term \textbf{N}ighttime \textbf{U}AV \textbf{T}racking benchmark with 42 sequences and 95,274 images.  
One-pass evaluation is adopted, and the performances are ranked by success rate, precision, and normalized precision. 
All the evaluations are complemented with the same platform with training.

\begin{table*}[!t]
  \centering
  \caption{Comparison of SAM-DA and other SOTA methods on DarkTrack2021 and NUT-L. With SAM, SAM-DA-Track can achieve better performance, especially when dealing with special challenges for nighttime UAV tracking. $\uparrow$ denotes higher is better. Best performances are highlighted in \textbf{bold}.}
  \vspace{-5pt}
    \resizebox{\linewidth}{!}{
    \begin{tabular}{c|ccc|ccc|ccc|ccc}
    \toprule
    \multirow{3}[1]{*}{Tracker} 
    & \multicolumn{6}{c|}{DarkTrack2021}&  \multicolumn{6}{c}{NUT-L} \\
    
    \cmidrule{2-13}
    &\multicolumn{3}{c|}{Illumination variation}&\multicolumn{3}{c|}{Low ambient intensity }
    &\multicolumn{3}{c|}{Illumination variation}& \multicolumn{3}{c}{Low ambient intensity} \\
        \cmidrule{2-13} &    AUC$\uparrow$  & $P_{Norm}$ $\uparrow$ & P$\uparrow$ & AUC$\uparrow$  & $P_{Norm}$ $\uparrow$ & P$\uparrow$&    AUC$\uparrow$  & $P_{Norm}$ $\uparrow$ & P$\uparrow$ & AUC$\uparrow$  & $P_{Norm}$ $\uparrow$ & P$\uparrow$  \\
    \midrule
    \textbf{SAM-DA-Track} & \textbf{0.451} & \textbf{0.524} & \textbf{0.593}   & \textbf{0.386} & \textbf{0.448} & \textbf{0.470} &\textbf{0.399} & \textbf{0.463} &\textbf{0.533}  & \textbf{0.448} & \textbf{0.511} & \textbf{0.572} \\
    Baseline~\cite{9879981} & 0.421 & 0.499 & 0.570 & 0.358& 0.422 & 0.446 
    &0.346 & 0.402& 0.463   &0.398 &0.461 & 0.515\\
          TCTrack++& 0.414 & 0.482 &0.561    & 0.332 &0.380 & 0.418  & 0.295 & 0.341 & 0.411  & 0.313& 0.376 & 0.431\\
          TCTrack& 0.390 & 0.464 & 0.538  &0.307 & 0.370 & 0.404 & 0.304 & 0.368 & 0.446   & 0.319 & 0.397& 0.450\\
          SiamBAN & 0.422 & 0.491 & 0.566   & 0.369 & 0.429 & 0.450 & 0.337 & 0.384 & 0.451  & 0.393 & 0.456 & 0.515 \\
    SiamRPN++ & 0.384 & 0.459 & 0.522& 0.293 & 0.336 & 0.369 & 0.305 & 0.347 & 0.406 & 0.344 & 0.398 & 0.448 \\
          SiamFC++& 0.410 & 0.482 & 0.544    & 0.270 & 0.316 & 0.336  & 0.307& 0.366 & 0.410   & 0.275 & 0.336 & 0.365\\
           SiamAPN++& 0.359 & 0.428 & 0.486    &0.253 & 0.303 & 0.296 & 0.296 & 0.347 & 0.407 & 0.303 & 0.351 & 0.380 \\
            SiamAPN& 0.304 & 0.372 & 0.420    &0.174& 0.240 & 0.232 &0.253 & 0.323 & 0.378  & 0.243 & 0.319 & 0.344 \\
            SiamFC+\_CI& 0.353 & 0.435 & 0.513    & 0.353 & 0.422 & 0.464 & 0.248 & 0.310 & 0.379  & 0.269 & 0.339 & 0.385 \\
            SiamFC+\_CR& 0.350 & 0.422 & 0.503   & 0.306 & 0.347 & 0.383 & 0.246& 0.300 &0.362   & 0.287 & 0.342 & 0.376 \\
            SiamFC+\_CX& 0.329& 0.395 & 0.471    & 0.290 & 0.330 & 0.353 & 0.239 & 0.290 & 0.351   & 0.254 & 0.305 & 0.347 \\
            SiamRPN+\_CR& 0.303 & 0.398 & 0.462    & 0.214 & 0.270 & 0.305 & 0.187 & 0.245 &0.291   & 0.164 & 0.219 & 0.221 \\
          UpdateNet& 0.340 &0.400 & 0.485    & 0.246 & 0.279 & 0.304  & 0.262 & 0.282 & 0.351  & 0.309 & 0.322 & 0.396\\
           Ocean& 0.392 & 0.472 & 0.529   & 0.380 & 0.434 & 0.448 & 0.237 & 0.284 & 0.318  & 0.238 & 0.286 & 0.313 \\
           SE-SiamFC& 0.313 & 0.385 &0.464   & 0.242 & 0.308 & 0.304 & 0.217 & 0.274 & 0.325   & 0.232 & 0.286 & 0.328 \\

    \bottomrule
    \end{tabular}
    }%
  \label{tab:attr}%
  
  \vspace{-5pt}
\end{table*}%

\begin{table*}
  \centering
  \caption{Comparison of SAM-DA and Baseline. With the fewer-better training, SAM-DA can achieve better performance on fewer raw images with more training samples and quicker training. Training duration is obtained on a single NVIDIA A100 GPU.
  Best performances are highlighted in \textbf{bold}. $\uparrow$ denotes higher is better while $\downarrow$ is the opposite.}
    \resizebox{\linewidth}{!}{
    \begin{tabular}{c|ccc|ccccc}
    \toprule
    \multirow{2}[4]{*}{Method} & \multirow{2}[4]{*}{Target domain dataset} & \multirow{2}[4]{*}{Images} & \multirow{2}[4]{*}{Data proportion} & \multirow{2}[4]{*}{Training samples$\uparrow$} & \multirow{2}[4]{*}{Traing duration$\downarrow$} & \multicolumn{3}{c}{NUT-L} \\
\cmidrule{7-9}          &       &       &       &       &       & AUC$\uparrow$  & $P_{Norm}$ $\uparrow$ & P$\uparrow$ \\
    \midrule
    Baseline~\cite{9879981} & NAT2021-$train$~\cite{9879981} & 276,081 & 100\% & 276,081 & 12h   & 0.377 & 0.434 & 0.498 \\
    \midrule
    \multirow{4}[2]{*}{SAM-DA} & {SAM-NAT-N} & 27,745 & 10.0\% & 1,608,843 & \textbf{2.4h} & 0.411 & 0.471 & 0.542 \\
          & {SAM-NAT-T} & 91,523 & 33.2\% & 5,314,760 & 4h    & 0.414 & 0.474 & 0.545 \\
          & {SAM-NAT-S} & 138,242 & 50.1\% & 8,042,926 & 6h    & 0.419 & 0.477 & 0.555 \\
          & {SAM-NAT-B} & 276,081 & 100\% & \textbf{16,073,740} & 12h   & \textbf{0.430} & \textbf{0.492} & \textbf{0.564}\\
    \bottomrule
    \end{tabular}
    }%
  \label{tab:all}%
\end{table*}%

\subsection{Overall evaluation}
This section provides a comprehensive analysis of trackers in nighttime UAV tracking with practical scenarios. The proposed tracker is based on the SAM-DA framework (dubbed as SAM-DA-Track). According to the version of training data, four trackers are acquired, namely SAM-DA-Track-B, SAM-DA-Track-S, SAM-DA-Track-T, and SAM-DA-Track-N. For fair comparison, SAM-DA-Track-B, dubbed as SAM-DA-Track and other 15 SOTA trackers~\cite{9879981, 9157457, 8954116, xu2020siamfc++, 8953458, 9008117, zhang2020ocean, 9636309, cao2022tctrack, fu2021siamese,cao2023towards} are overall evaluated on DarkTrack2021 and the proposed NUT-L.
As shown in Fig.~\ref{fig:overall} (a), SAM-DA-Track promotes Baseline on DarkTrack2021 by \textbf{7.8\%}, \textbf{5.8\%}, and \textbf{5.0\%} on success rate, normalized precision, and precision, respectively. On NUT-L, Fig.~\ref{fig:overall} (b) shows the proposed SAM-DA-Track ranks first by a large margin compared to other trackers. Specifically, SAM-DA-Track raises Baseline~\cite{9879981} by \textbf{14.1\%}, \textbf{13.4\%}, and \textbf{13.3\%} on three metrics. The results demonstrate that SAM-DA-Track presents better adaptability and practicality in variant nighttime conditions. The improvement is attributed to the enormous high-quality training samples powered by the superior zero-shot generalization ability and robustness of SAM. 

\subsection{Attribute-based performance}
Objects in actual nighttime UAV tracking scenarios usually experience complex and varied lighting issues. This mainly includes two aspects, the target object undergoes drastic changes in illumination (illumination variation, IV) and the object is under extremely low light conditions (low ambient intensity, LAI). Table~\ref{tab:attr} presents the performance of SAM-DA-Track and other SOTA trackers against two challenges. Compared with the Baseline on IV, SAM-DA-Track achieves a 7.1\%, 5.0\%, 4.0\%, 15.3\%, 15.2\%, and 15.1\% improvement in the three metrics on DarkTrack2021 and NUT-L, respectively. While on LAI, SAM-DA-Track promotes the Baseline by 7.8\%, 6.2\%, 5.4\%, 12.6\%, 10.8\%, and 11.1\%. 
The evaluation of the lighting challenges has verified the robustness of SAM-DA-Track against the severe issues for practical nighttime UAV tracking.

\begin{figure}[t!]	
	\centering
{
	\includegraphics[width=0.98\linewidth]{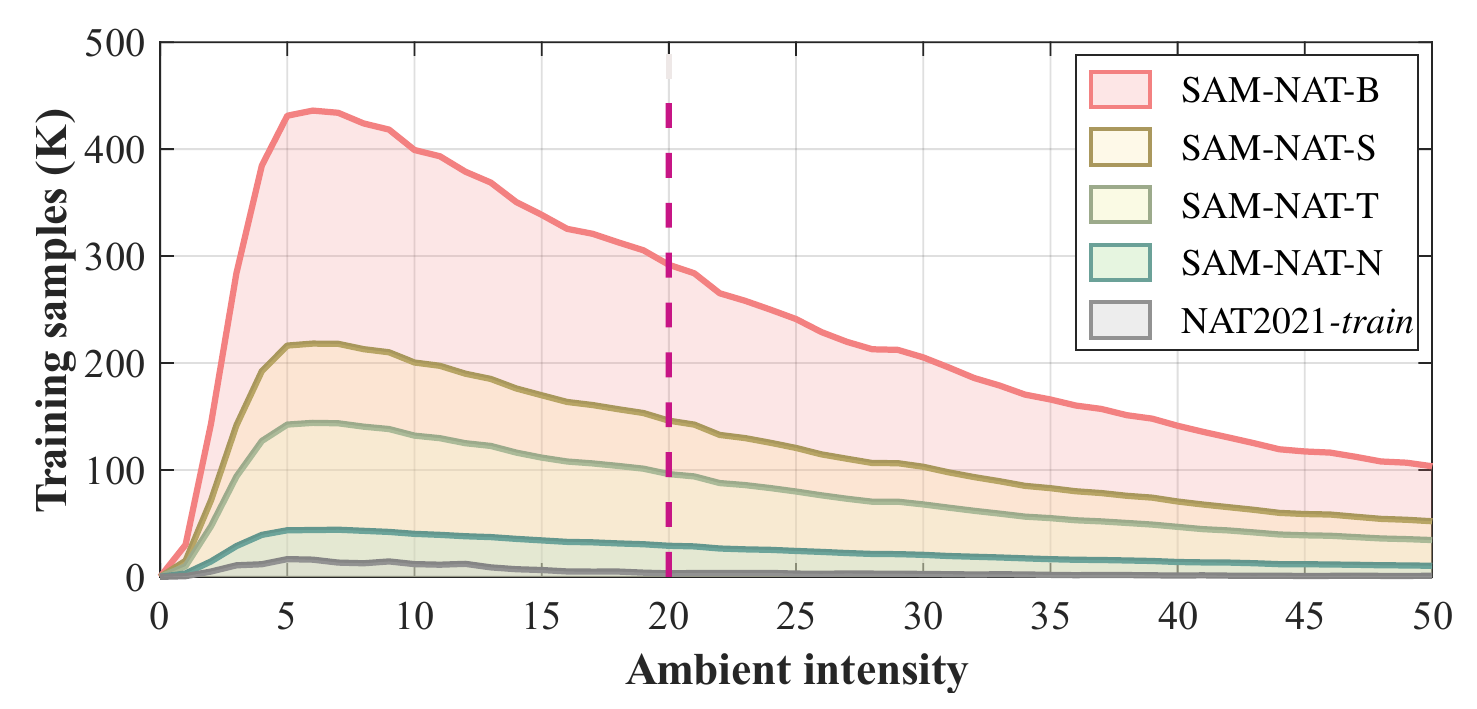}
}
	\caption
	{
        Ambient intensity (AI) distribution comparison between four versions of {SAM-NAT} and NAT2021-$train$. Enormous training samples with diverse lighting conditions are enriched with SAM-powered target domain training sample swelling, especially those with AI values less than 20 (the \textcolor[rgb]{0.78,0.08,0.52}{red} line), which are seen as low ambient intensity in~\cite{9879981}.
	}
	\label{fig:ai}
\end{figure}

\Remark As a foundation model for segmentation, SAM presents its superior zero-shot performance even in extremely dark nighttime images, as shown in Fig.~\ref{fig:ai}. Furthermore, this encouraging phenomenon enables more powerful domain adaptation in other downstream tasks against specific challenges of nighttime domain data.

\subsection{Analysis on training sample swelling}\label{sec:labelswelling}
Considering the scarcity of nighttime tracking data and the high cost of annotating data under unfavorable lighting conditions, the efficiency of utilizing existing nighttime data is crucial.
The core idea of SAM-DA is to utilize the strong zero-shot generalization ability and robustness of SAM to produce enormous high-quality training samples in the nighttime tracking data. The high quality of swelled training samples is shown in Fig.~\ref{fig:visual}, where SAM can automatically produce pixel-level determined masks with clear boundaries even in the low-light environment. Notably, no low-light image enhancement is required, which is different from the Baseline. Besides, the number of training samples is greatly enriched, which represents the diversity of swelled training samples.
Compared with~\cite{9369102} and~\cite{zhang2021dynamic} used in Baseline, SAM can discover anything potential for tracking. In addition to common objects used in nighttime UAV tracking like cars and people, SAM also includes other valuable tracking candidates, \textit{e.g.}, monitors, and traffic signs in Fig.~\ref{fig:visual}. Therefore, nighttime tracking benefits from the generalization ability and robustness of SAM against complicated scenes.

\noindent{\textbf{Enlarged training samples.}} As shown in Tab.~\ref{tab:all}, the NAT2021-$train$ includes 276,081 training samples in 276,081 training images, with only a single object in each image. Many potential objects remain undiscovered. By contrast, {SAM-NAT-N} contains 1,608,843 training samples with only 10.0\% of training images in NAT2021-$train$. The number of training samples of {SAM-NAT-N} is already \textbf{5.8} times of NAT2021-$train$. 
Besides, {SAM-NAT-T} uses 33.2\% of images and reaches 5,314,760 training samples, while {SAM-NAT-S} includes 50.1\% of images and contains 8,042,926 training samples.
{SAM-NAT-B} uses equal amounts of images with NAT2021-$train$ and contains 16,073,740 training samples, which astonishingly reaches \textbf{58.2} times compared to NAT2021-$train$.

\noindent{\textbf{Enriched lighting conditions.}} Figure~\ref{fig:ai} demonstrates the ambient intensity (AI) comparison between {SAM-NAT} and {NAT2021-$train$}. The AI value is calculated based on the average lighting conditions of the image patches, where the lower the AI value, the darker the ambient environment in which the target object is located. The patches with AI value of less than 20 are regarded with the attribute of low ambient intensity~\cite{9879981}. Diverse lighting conditions in NAT2021-$train$ are all enriched in {SAM-NAT}, especially for the training samples with AI value less than 20, \textit{i.e.}, low ambient intensity. The comparison validates that the SAM-powered target domain training sample swelling can enrich the distinguished characteristics of the target domain (low light conditions in this case), thus improving the knowledge transfer ability of domain adaptation.

\begin{figure}[!t]	
 \vspace{-10pt}
	\centering
{
	\includegraphics[width=0.96\linewidth]{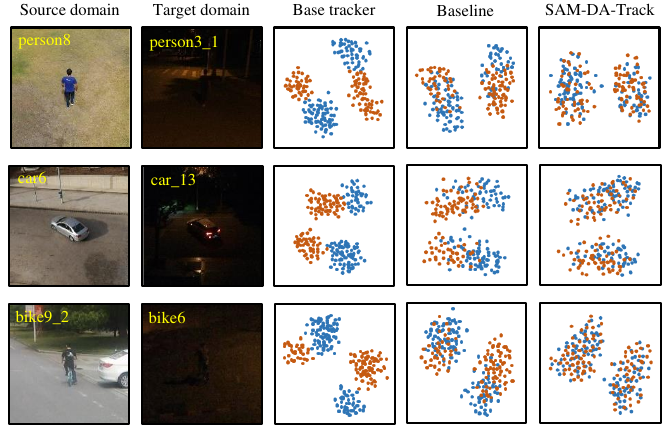}
 \vspace{-5pt}
	\caption
	{
		Feature visualization by t-SNE \cite{van2008visualizing} of day-night similar scenes. The daytime frames are from UAV123 \cite{mueller2016benchmark} and UAVTrack112 \cite{9477413}, while the nighttime frames are from NUT-L. \textcolor[rgb]{0.18,0.46,0.71}{Blue} and \textcolor[rgb]{0.77,0.35,0.07}{orange} indicate the source and target domain, respectively. The scattergrams depict day-night features from the base tracker, Baseline, and SAM-DA-Track. The results show that SAM-DA achieves superior performance to narrow domain discrepancy.
	}
	\label{tsne}
 }
 \vspace{-5pt}
\end{figure}

\subsection{Fewer-better training}
Since SAM-DA provides better training samples with fewer data, an intriguing topic is to discuss \textit{can tracker achieve better performance with less training}. 
This is highly relevant to practical nighttime UAV applications, where the amount of training data is usually limited, and quick training is required for timely implementation. The results in Tab.~\ref{tab:all} validate that even with very constrained training image proportion (10.0\% on SAM-NAT-N) and training time (about 2.4 hours), SAM-DA-Track can achieve better performance (0.411) than Baseline (0.378) on NUT-L. It proves the practicality of higher training efficiency with less data. With more training data and longer training time, SAM-DA-Track achieves further improvement. With swelling on whole data ({SAM-NAT-B}) and the same training time as Baseline, SAM-DA obtains further improvement (0.430). The fewer-better training evaluation further validates the effectiveness of SAM-powered training sample swelling. It also demonstrates the proposed method is not data-hungry with enormous high-quality training samples.
\begin{figure}[!t]
 
	\centering	
		\includegraphics[width=0.96\linewidth]{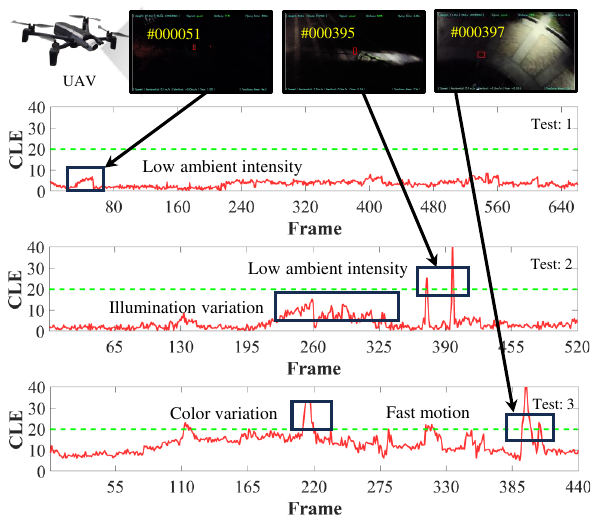}
	\caption
	{
  		Several results of real-world tests. The CLE below the \textcolor[rgb]{0,1,0}{green} dashed
the line is the success tracking in the real-world test.
	}
	\label{fig:real}
\end{figure}

\subsection{Day-night feature distribution}
To validate the effectiveness and the domain adaptability of the proposed method, this section includes visualizations of the day-night image features obtained from the base tracker, the Baseline, and the SAM-DA-Track. Figure \ref{tsne} shows the visualization results using t-SNE~\cite{van2008visualizing}. Comparing the proposed SAM-DA with the base tracker and the Baseline, SAM-DA has further reduced the domain discrepancy significantly by leveraging enormous high-quality target domain training samples. As a data-driven method with remarkable zero-shot generalization ability, SAM has been proven that it can be applied for enlarging the target domain training samples for tracking-oriented day-night domain adaptation.

\subsection{Real-world tests}
To demonstrate the applicability of SAM-DA-Track in nighttime UAV tracking, numerous real-world tests are implemented. A laptop carrying an NVIDIA RTX 3060 GPU serves as the ground control station (GCS). The Parrot UAV captures the image frames and transmits them to the GCS through WiFi communication. The tracker predicts the bounding box with 32 frames per second (fps) and then sends it to the Parrot UAV. The performance of three typical scenes is demonstrated in Fig.~\ref{fig:real}. The CLE curves represent the error between the estimated location and ground truth. In the tests, SAM-DA-Track perform well in real-time when confronted with fast motion, low resolution, occlusion, and illumination variation. 

\section{Conclusion}
This work is the first study to introduce the superior SAM into the training phase of day-night domain adaptation for nighttime UAV tracking, proposing a novel SAM-powered domain adaptation framework, \textit{i.e.}, SAM-DA.
Specifically, the SAM-powered target domain training sample swelling is designed to determine enormous high-quality target domain training samples from every single challenging nighttime image. The above one-to-many generation significantly increases the high-quality target domain training samples for day-night domain adaptation. Consequently, the reliance on the number of raw images can be decreased, enhancing generalization and preventing overfitting. Extensive evaluation on enormous nighttime videos shows the robustness and domain adaptability of SAM-DA for nighttime UAV tracking. To summarize, this work can contribute to the advancement of domain adaptation for object tracking and other vision tasks in various unmanned systems.

\bibliographystyle{IEEEtran}
\balance
\bibliography{main}

\end{document}